\documentclass[letterpaper]{article} % DO NOT CHANGE THIS
\usepackage{aaai25}  % DO NOT CHANGE THIS
\usepackage{times}  % DO NOT CHANGE THIS
\usepackage{helvet}  % DO NOT CHANGE THIS
\usepackage{courier}  % DO NOT CHANGE THIS
\usepackage[hyphens]{url}  % DO NOT CHANGE THIS
\usepackage{graphicx} % DO NOT CHANGE THIS
\usepackage{natbib}  % DO NOT CHANGE THIS AND DO NOT ADD ANY OPTIONS TO IT
\usepackage{caption} % DO NOT CHANGE THIS AND DO NOT ADD ANY OPTIONS TO IT
\usepackage{amsmath,amssymb,amsfonts}
\usepackage[linesnumbered,ruled,vlined]{algorithm2e}
\SetKw{Continue}{continue}

\usepackage{hyperref} %-- This package is specifically forbidden

\title{Safe Interval Randomized Path Planning For Manipulators}

\author {
    Nuraddin Kerimov\textsuperscript{\rm 1, 2},
    Aleksandr Onegin\textsuperscript{\rm 2},
    Konstantin Yakovlev\textsuperscript{\rm 1, 3}
}
\affiliations {
    \textsuperscript{\rm 1}FRC CSC RAS~~\textsuperscript{\rm 2}MIPT~~\textsuperscript{\rm 3}AIRI\\
    \textit{Correspondence to: kerimov.nm@phystech.edu}
}

\begin{document}

\maketitle

\begin{abstract}
Planning safe paths in 3D workspace for high DoF robotic systems, such as manipulators, is a challenging problem, especially when the environment is populated with the dynamic obstacles that need to be avoided. In this case the time dimension should be taken into account that further increases the complexity of planning. To mitigate this issue we suggest to combine safe-interval path planning (a prominent technique in heuristic search) with the randomized planning, specifically, with the bidirectional rapidly-exploring random trees (RRT-Connect) -- a fast and efficient algorithm for high-dimensional planning. Leveraging a dedicated technique of fast computation of the safe intervals we end up with an efficient planner dubbed SI-RRT. We compare it with the state of the art and show that SI-RRT consistently outperforms the competitors both in runtime and solution cost.

Our implementation of SI-RRT is publicly available at \\
\url{https://github.com/PathPlanning/ManipulationPlanning-SI-RRT}

\end{abstract}

\section{Introduction}
Robotic manipulators are the high degrees-of-freedom (DoF) robotic systems that are traditionally used in industrial applications where their movements can be precomputed. Meanwhile, in numerous other settings their paths should be planned adaptively taking the current state of the environment into account -- think, for example, of a household mobile robot that is equipped with a manipulator and uses it for various pick'n'place tasks.

Generally, sampling-based planning algorithms, such as RRT~\cite{lavalle1998rapidly}, PRM~\cite{kavraki1996probabilistic} and their numerous modifications, are widely used for manipulation path planning as these methods (in contrast to the search-based ones rooted in A*~\cite{hart1968formal} algorithm) can better handle the high dimensionality of the search space (as conventional modern manipulators typically have 5, 6, 7 DoF). 

In this work we are specifically interested in path planning for high DoF manipulators in dynamic environments, i.e. the ones there the moving obstacles are present. If the obstacles' movements can not be predicted, reactive approaches based on fast re-planning can be used to solve the problem. Such sampling-based methods as RRT$^X$~\cite{otte2016rrtx} or DRGBT~\cite{covic2021path} can be named as the representatives of such solvers. Fast search based re-planning methods are also known, with D*Lite~\cite{koenig2002d} being one of the most widespread solver of this kind (especially in mobile robotics).

In this work, however, we assume that the trajectories of the moving obstacles are known. E.g. they are accurately predicted via an external motion prediction module or these obstacles are actually the other robotic systems that execute the known trajectories. In this case it is beneficial to take the information on the obstacles' movements into account while planning. To this end, a planner should reason about the time dimension as well. An example how this reasoning can be applied to sampling-based planning is presented in~\cite{sintov2014time}, where the arrival time is fixed beforehand and time is added to the nodes in the search tree such that each node denotes a specific state in a specific time. A more recent and advanced method is ST-RRT*~\cite{grothe2022st}. It is a bidirectional search algorithm that does not need to specify the arrival time but rather updates it on-the-fly while searching. Currently, ST-RRT* can be deemed as state of the art in sampling based planning with time dimension.

\begin{figure}[t]
    \centering
    \includegraphics[width=0.35\textwidth]{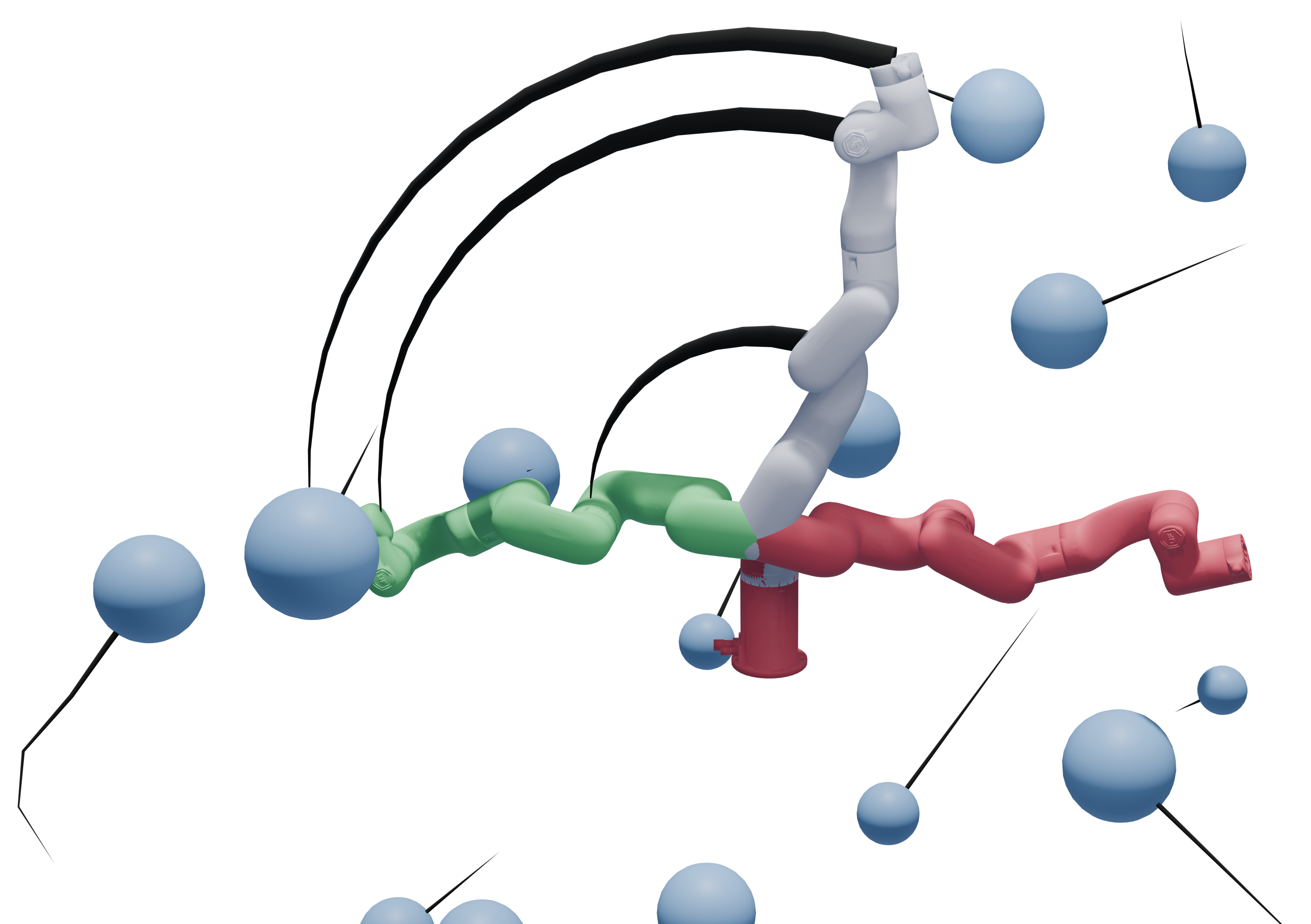} 
    \caption{A problem we are interested in -- path planning for a manipulator in the presence of moving obstacles whose trajectories are known (accurately predicted).
    %Manipulator on a synthetic stage with randomly flying spherical obstacles. Green indicates the start configuration of the robot, white is the intermediate configuration on the constructed path, red is the goal configuration. Blue spheres are obstacles. Black lines are motion trails.
    }
    \label{fig:example}
\end{figure}

In this short paper we wish to advance this state of the art by incorporating the idea of safe interval path planning (SIPP) into the (bidirectional) sampling-based search framework. SIPP was originally introduced in~\cite{phillips2011sipp} as a search-based planner. However the main idea of SIPP, i.e. reasoning not over the distinct time moments but rather over the time intervals, is applicable to sampling-based planning as well. Our planner that utilizes this idea, SI-RRT, is shown to notably outperform the competitors in both runtime and solution quality. It can successfully solve the instances with a very high number of moving obstacles where the competitors fail.

We note that we are not the first to explore the combination of safe interval path planning and sampling-based planning. E.g. in~\cite{li2019safe} a combination of SIPP and PRM was proposed for manipulation planning and in~\cite{sim2024safe} a combination of SIPP and RRT* was presented for low-dimensional path planning, i.e. to 2D pathfinding. Still, to the best of our knowledge, we are the first to combine SIPP and RRT for high DoF planning, utilize bidirectional search and introduce an dedicated technique to estimate safe intervals that is a key to high efficiency of our solver.

\section{Problem statement}
Let $\mathcal{W} \subset \mathbb{R}^3$ be the workspace of a manipulator consisting of $n$ joints. Its configuration space,  $\mathcal{C}$, is composed of tuples $\textbf{q} = \{q_1, ..., q_n\}$, where $q_i$ is the rotation angle of the $i$-th joint with respect to the previous one. $T = [0, t_{\text{max}}]$ is time, where $t_{\text{max}}$ is the maximum time allowed for the operation.

Besides manipulator $K$ obstacles move through the environment and their trajectories are known. Let $\mathcal{O}_i(t)$ denote the subset of $\mathcal{W}$ occupied by the $i$-th obstacle at time $t$; and $\mathcal{O}(t)=\bigcup_{i=1}^K\mathcal{O}_i(t)$.

A (timed) path or a trajectory of a manipulator is a mapping $\pi(t): [0,t_{arrival}] \rightarrow \mathcal{C}$. A path is collision free if $\forall t \in [0, t_{arrival}]: \mathcal{R}(\pi(t)) \cap \mathcal{O}(t) = \emptyset$, where $\mathcal{R}(\pi(t))$ denotes a subset of $\mathcal{W}$ occupied by robot at time $t$ according to the path $\pi$.

The problem is as follows. Given the start and the goal configurations, $\textbf{q}_{start}$ and $\textbf{q}_{goal}$ find $t_{arrival} \leq t_{max}$ and a collision-free path, $\pi(t)$, for a manipulator, s.t. $\pi(0)=\textbf{q}_{start}$ and $\pi(t_{arrival})=\textbf{q}_{goal}$.

%Suppose we have a function $h(\textbf{q}, t) \in \mathcal{W} \times \mathcal{T}$ that returns by the configuration $\textbf{q}$ all points of the robot in the workspace at time $t$. Let $(\mathcal{C} \times \mathcal{T})_{\text{free}} = \{ (\textbf{q}, t) \ | \  h(\textbf{q}, t) \cap \mathcal{O}(t) = \emptyset \}$ be space-time free configuration space. The aim of our problem is to find a trajectory $\pi(t) : [0, t^{\text{arrival}}] \to (\mathcal{C} \times \mathcal{T})_{\text{free}}$ for a given initial and final configuration of the robot, $\textbf{q}_{\text{start}}$ and $\textbf{q}_{\text{goal}}$. We also assume that the robot can either move at its maximum velocity $v_{max}$ or stand still. Let $GSI(\textbf{q})$ be a function that gets a set of safe intervals, $si = [t_{\text{low}}, t_{\text{high}}]$ for configuration $\textbf{q}$. A safe interval is a continuous time period during which a particular configuration of a robot remains collision-free with respect to moving obstacles in the environment.

\section{Method}

The idea of our method, SI-RRT, is to combine in a single framework RRT-Connect, a single-query sampling-based planner utilizing bidirectional search, and SIPP -- a search-based planner in space-time that reasons over the time intervals instead of distinct time moments. We also rely on the following assumptions. First, we assume that the inertial effects are neglected and the manipulator can either stay put or move with the constant velocity. Second, we rely on an auxiliary collision checking routine that given the configuration of the robot and positions of the obstacles checks whether the latter is collision free. Indeed, this procedure may be computationally burdensome.

SI-RRT grows the two search trees, one rooted in the start configuration, $\mathcal{T}_{start}$, and the other one rooted in the goal configuration, $\mathcal{T}_{goal}$, until they meet. A node of a tree is identified by a tuple $n=(\textbf{q}, si=[t_l, t_u])$, where $si$ is a \emph{safe interval} -- a continuos time interval comprised of the time moments when the robot configured at $\textbf{q}$ does not collide with the obstacles. We will elaborate on how to compute $si$ given $\textbf{q}$ later on. 

Each node $n$ is characterized by the arrival time $time(n) \in si$ and its predecessor -- $parent(n)$. For the nodes residing in the start tree, $time(n)$ is the earliest possible time a robot can reach $n$ from $parent(n)$. Ideally, we want to perform $parent(n) \rightarrow n$ movement immediately after arriving $parent(n)$. However committing to such move immediately may result in collision. To this end we utilize the so-called `wait-and-go' actions introduced in the original SIPP paper. That is, we wait at $parent(n)$ for the minimum possible time so that the moving action becomes safe and then perform it. For the goal tree, $\mathcal{T}_{goal}$, the reversed reasoning is applied. I.e. we want to arrive as late as possible to the successor and $time(n)$ is computed accordingly (it is the latest possible arrival time).

We will now elaborate in more details on how SI-RRT grows its trees building upon the pseudocode presented in Alg.~\ref{alg:sirrt}. We initialise $\mathcal{T}_{start}$ with the node comprised of $\textbf{q}_{start}$ and the first safe interval of this configuration (we assume that it starts with 0). The arrival time is set to $0$. $\mathcal{T}_{goal}$ is initialised by the node comprised of $\textbf{q}_{goal}$ and the latest safe interval of this configuration (the one that contains $t_{max}$). Arrival time is set to $t_{max}$. Then we initialise $\mathcal{T}_{current}$ as $\mathcal{T}_{start}$ and $\mathcal{T}_{other}$ as $\mathcal{T}_{goal}$.

\begin{algorithm}[tb]
\caption{SI-RRT Planner}
\label{alg:sirrt}
% \SetAlgoLined
% \DontPrintSemicolon
\KwIn{$\textbf{q}_{start}$, $\textbf{q}_{goal}$, $t_{max}$, $\Delta_{planner}$, $\Delta_{parent}$}

$n_{start} := (\textbf{q}_{start}, si_{first})$; $time(n_{start})=0$\;
$n_{goal} := (\textbf{q}_{goal}, si_{last})$; $time(n_{goal})=t_{max}$\;

$\mathcal{T}_{start} := \{n_{start}\}$; $\mathcal{T}_{goal} := \{n_{goal}\}$\;

$\mathcal{T}_{current} := \mathcal{T}_{start}$; $\mathcal{T}_{other} := \mathcal{T}_{goal}$\;

\While{goal not reached or have time to plan}{
    $\textbf{q}_{sampled} :=$ \texttt{sampleCFG()}\;
    
    $\textbf{q}_{new} := $ \texttt{extend}($\textbf{q}_{sampled}$, $\mathcal{T}_{current})$\;

    \If{$\textbf{q}_{new} = \emptyset$}{
        \texttt{swap}($\mathcal{T}_{current}$, $\mathcal{T}_{other}$)\;
        \Continue
    }
    $\{si\} := \texttt{getSafeIntervals}(\textbf{q}_{new})$\;

    $\{n_{new}\} := \texttt{setParent}(\mathcal{T}_{current}, \textbf{q}_{new}, \{si\})$\;
    
    \If{$\{n_{new}\} = \emptyset$}{
        \texttt{swap}($\mathcal{T}_{current}$, $\mathcal{T}_{other}$)\;
        \Continue
    }

    $\mathcal{T}_{current}.Add(\{n_{new}\})$\;

    $\{n_{join}\}:= \texttt{connect}(\mathcal{T}_{other}, \textbf{q}_{new}, \{n_{new}\})$\;

    \If{$\{n_{join}\} \ne \emptyset$}{
        $\texttt{uniteTrees}(\mathcal{T}_{start}, \mathcal{T}_{goal}, \{n_{join}\})$\;
        \Return success\;
        
        }
    
    \texttt{swap}($\mathcal{T}_{current}$, $\mathcal{T}_{other}$)\;
}

\Return fail\;
\end{algorithm}

%We named our planner SI-RRT. Pseudocode of SI-RRT is listed in Algorithm~\ref{alg:sirrt}. It uses two trees, start tree $\mathcal{T}_{start}$ and goal tree $\mathcal{T}_{goal}$. Nodes of each tree $n$ are defined by set of robot configuration $\textbf{q}$, safe interval $si$, which is described by lower and upper time bound, the time moment $t_{\text{arrival}}$ when robot arrives in safe interval $si$ at $\textbf{q}$, pointer to parent node $n_{parent}$, the time moment $t_{\text{departure}}$ when robot departures from parent's safe interval. We initialise $\mathcal{T}_{start}$ by node in $\textbf{q}_{start}$ and earliest safe interval at $\textbf{q}_{start}$, arrival time is set to $0$, departure time is set to $-1$.  $\mathcal{T}_{goal}$ is initialised by node in $\textbf{q}_{goal}$ and latest safe interval at $\textbf{q}_{goal}$, arrival time is set to $t_{max}$, and departure time is set to -1. We assume, that the latest safe interval at $\textbf{q}_{goal}$ will have upper bound time at $t_{max}$. For the $\mathcal{T}_{start}$ we try to achieve new nodes as early as possible to ensure lowest arrive time and better reachability for future nodes. For the $\mathcal{T}_{goal}$ with each node we  travel back in time with respect to $v_{max}$, and we try to reach new nodes as late as possible to ensure reachability. Then we initialise $\mathcal{T}_{current}$ as $\mathcal{T}_{start}$ and $\mathcal{T}_{other}$ as $\mathcal{T}_{goal}$.

After initialising the trees, at each iteration of the algorithm we randomly sample a new configuration $\textbf{q}_{sampled}$ with the \texttt{sampleCFG} function. Then, in the $\texttt{extend}(\mathbf{q}, \mathcal{T}_{\text{current}})$ function, we first compute the nearest configuration, $\mathbf{q}_{\text{near}}$, to the sampled configuration, $\mathbf{q}_{\text{sampled}}$, in the current tree, $\mathcal{T}_{\text{current}}$, using the $L_2$ distance. Subsequently, a new configuration $\mathbf{q}_{\text{new}}$ is computed by advancing from $\mathbf{q}_{\text{near}}$ in the direction of $\mathbf{q}_{\text{sampled}}$ by a distance limited to the maximum step size, $\Delta_{\text{planner}}$, which is an input parameter. If the configuration is not valid, i.e. robot is in collision with the static obstacles, we return $\emptyset$, swap the trees and skip to the next iteration. Otherwise, $\mathbf{q}_{\text{new}}$ is returned from the function and next we compute the safe intervals for it.

Computing the safe intervals is done by using the auxiliary collision detection routine, that takes as input the robot's configuration (and shape) and the positions (and shapes) of the obstacles and outputs a boolean indicating whether the collision exists or no. This routine should be sequentially executed at consecutive time steps to detect which ones result in collision and which ones do not. The latter time moments will form the safe intervals of the configuration. In practice the frequency of the collision checks should be high enough not to miss the collisions. E.g. in out tests we set this frequency to 30 times per second, while $t_{max}$ was set to 20s. This infers that to compute the safe intervals for any configuration we need 600 collision checks, which becomes computationally expensive.

To mitigate this issue we rely on the specifics of the collision checking procedure. Typically for manipulation planning collision detection is composed of two stages: the broad phase, when possible colliding obstacles are identified, and the narrow phase, when the detailed collision checks are carried out and the answer is provided. To speed up the process we create a scene on which we put all the obstacles at the positions corresponding to all the time steps and we keep the information on which obstacle comes from which time step. Then we perform a single broad phase collision checking that identifies the obstacles (annotated with the time steps) that need to be passed further to the narrow collision detection. Thus we safe time not invoking broad collision detection sequentially.

%. We make an assumption, that obstacle movements are discrete in time and updated each frame, and amount of frames per second is constant. Naive way to get safe intervals is iteratively check collision through frames with fixed manipulator configuration. Our approach is to join all obstacles at one frame, but leave information about their frames into them. Then, instead of iterating through frames, we find all collisions in one unified frame. This speeds up algorithm by 10 times, because, instead of broad phase-narrow phase collision detection at each frame, only one broad phase collision detection is made, thus, this method effectively finds candidates for narrow phase collision detection.

After the safe intervals are computed, in $\texttt{setParent}$ function we try to connect each safe interval, $si$, with one the nearest nodes lying no further away than $\Delta_{\text{parent}}$ from $\mathbf{q}_{new}$. As detailed before, for the $\mathcal{T}_{start}$ we try to arrive to a safe interval as early as possible, and for the $\mathcal{T}_{goal}$ we try to arrive to a safe interval as late as possible. If we succeed we add the new node $n_{new}=(\mathbf{q}_{new}, si)$ to a set of new nodes $\{n_{new}\}$ that are added further to the tree.

%, travelling back in time. First, we sort candidate nodes by their arrival time. We make an assumption, that from the earliest (latest in case of $\mathcal{T}_{goal}$) arrival time of parent node we can get the earliest (latest) arrival time of new node. For each safe interval from which we can reach the $\textbf{q}_{goal}$ with respect to $t_{max}$  and $v_{max}$, we go through all possible departure times and find one, where, when travelling at maximum speed, we reach new configuration within the safe interval without collision. After all of that, we return list of new nodes $\textbf{n}_{new}$, that were created within $\textbf{q}_{new}$. 

%If new nodes were successfully created, function $\texttt{connect}(\mathcal{T}_{other},q_{new},\{n_{new}\})$ is called. It attempts to greedily extend $\mathcal{T}_{other}$ in order to connect it to $\mathcal{T}_{current}$. In specifics if identifies the node of $\mathcal{T}_{other}$ which is the closest to $\textbf{q}_new$ and sequentially

\begin{figure}[t]
    \centering
    \includegraphics[width=0.32\textwidth]{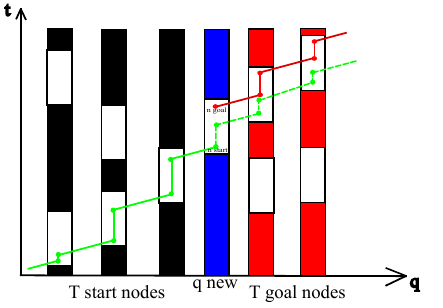} 
    \caption{Trimming the wait action in a constructed path.}
    \label{fig:trim}
\end{figure}

If new nodes were successfully created, function $\texttt{connect}(\mathcal{T}_{other},q_{new},\{n_{new}\})$ is called in which we try to connect $\mathcal{T}_{other}$ with $\mathcal{T}_{current}$ by greedily extending the latter towards $\textbf{q}_{new}$. That is we iteratively append new nodes to $\mathcal{T}_{other}$ until we either reach $\textbf{q}_{new}$  or we can not extend further (due to collisions with the obstacles). In the former case we have two branches in $\mathcal{T}_{current}$ and $\mathcal{T}_{other}$, that meet at $\textbf{q}_{new}$ at a certain safe interval. We then invoke \texttt{uniteTrees} function that concatenates these branches and reverses the edges of the branch that belongs to $\mathcal{T}_{goal}$ so all actions now go forward in time. These actions constitute the sought path. Thus we finish the algorithm with the success. 

Please note, that the found path may contain a prolonged wait action in $\textbf{q}_{new}$ as the forward branch has been build with the intention of reaching it as early as possible while the backward branch -- as late as possible. We can appropriately trim this wait action (i.e. decrease the wait to the minimal possible duration s.t. the chain of actions outgoing from $\textbf{q}_{new}$ is still consistent with the respective safe intervals) -- the illustration of this process is provided in Fig.~\ref{fig:trim}. In our experiments we use this trimming.

\section{Experiments}

We evaluate SI-RRT\footnote{Our C++ implementation of SI-RRT is available at \\ \url{https://github.com/PathPlanning/ManipulationPlanning-SI-RRT}} and compare it with state-of-the-art competitors in 3D path planning for 6 DoF manipulator, Ufactory xArm 6, modelled as a chain of 3D capsules for collision checking. Its maximal length is approx. $0.7$m. 

For the evaluation we create a wide range of path planning instances that involve different number of moving obstacles, from 1 to 300. Each moving obstacle is modelled as a sphere of radius $r$ that is randomly sampled from the interval $[0.05, 0.10]$m. It moves with the constant velocity randomly sampled from the interval $[0, 1]$m/s, changing the movement direction sporadically. When generating each problem instance we ensure that the obstacles do not hit the initial and the target configuration of the manipulator. An example of a problem instance is shown in Fig.~\ref{fig:example}. The animated visualization is available in the supplementary material.

For each number of obstacles we generate 50 different problem instances. The complexity of the instances increases with the number of obstacles. That is, we keep adding the obstacles to the previously generated ones with the increment of 20 (up to 300 obstacles per instance).

The time range in all tests, $t_{max}$, is set to $20$s. Each test is repeated 10 times. We compare our planner with ST-RRT*~\cite{grothe2022st}, a state-of-the-art solver that takes the trajectories of the obstacles into account and plans in space-time (like our planner); and with DRGBT~\cite{covic2021path} a modern and fast re-planning solver. We take the authors implemenation of ST-RRT*~\cite{ompllib}  and DRGBT~\cite{rpmpllib}. Both of these planners require setting several hyperparameters that influence their performance significantly. To this end, we conduct a grid search to identify the best values. %Noteworthy that for DRGBT we do not take $t_{max}$ into account, i.e. the solver reactively re-plans until either the final configuraition is reached or collision occurs. For ST-RRT* we utilize $t_{max}$ and halt the planner when the first solution is found and skip the path enhancement phase (in our setup we observed that this phase does not lead to non-marginal impovements while consuming time). 
For SI-RRT we use following hyperparameters: $\Delta_{planner} = 1$ radians, $\Delta_{parent} = 3$ radians.

\begin{figure}[t]
    \centering
    \includegraphics[width=\linewidth]{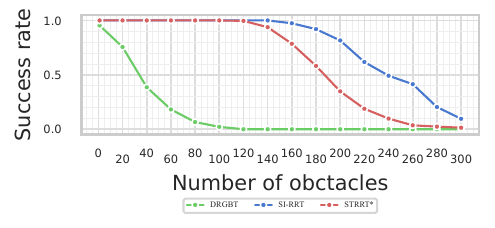} 
    \caption{Success rate of the evaluated planners.}
    \label{fig:success_rate}
\end{figure}

\begin{figure}[t]
    \centering
    \includegraphics[width=\linewidth]{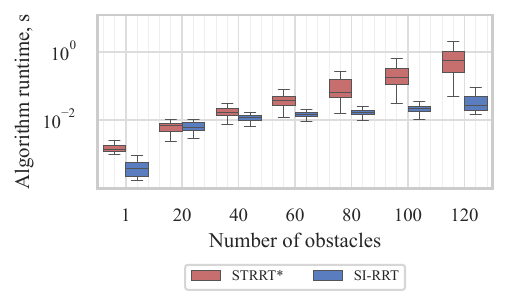} 
    \caption{Runtime of SI-RRT and ST-RRT* on the instances that were successfully solved by both of them.
    %The plot shows the operating time of  ST-RRT*(red) and SI-RRT(blue). The scale of the graph is limited to 120 obstacles, where both planners have success rate equal to 100\%. SI-RRT has much better time-limit asymptotics than ST-RRT*, overtaking it by an order of magnitude in runtime.
    }
    \label{fig:runtime}
\end{figure}

\begin{figure}[t]
    \centering
    \includegraphics[width=\linewidth]{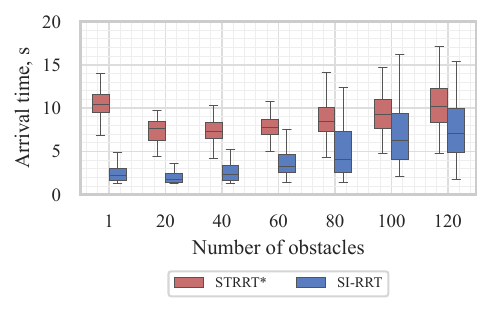} 
    \caption{Solution quality (time when the manipulator arrives at the goal configuration) of ST-RRT* and SI-RRT.
    %The plot shows the robot arrival time at the target for the ST-RRT*(red) and SI-RRT(blue) algorithms. ST-RRT* works until the first path is found, without further optimisation. Our SI-RRT method has much lower arrival time than ST-RRT*.
    }
    \label{fig:arrival}
\end{figure}

The results of the experiments are shown in Figures~\ref{fig:success_rate},~\ref{fig:runtime},~\ref{fig:arrival}. Specifically, Figure~\ref{fig:success_rate} shows the success rate (SR), i.e. the percentage of tasks that were successfully solved for the increasing number of obstacles. To count SR for DRGBT we consider a test as failed if the planner was not able to reach the goal until $t=t_{max}$ or the robot got in collision before. For ST-RRT* and SI-RRT we consider the test as failed if no solution was constructed under a time cap of 20s. Clearly, our planner outperforms the competitors. E.g. for 200 obstacles the SR of SI-RRT is 77\% while the one of ST-RRT* (the closest competitor) is 42\%.

%depending on the complexity of the task, i.e. the number of dynamic obstacles. The reactive real-time scheduler DRGBT has no runtime limit, i.e., it runs until it exceeds $T_{\text{max}}$ equal to 20 seconds or encounters an obstacle. DRGBT has a sharp drop in success rate from a small number of obstacles and reaches about zero at 120 obstacles. Deliberative ST-RRT* and SI-RRT have a runtime limit of 20 seconds. State-of-The-Art ST-RRT* has a 100\% success rate up to 120 obstacles and then gradual drop in success rate occurs. SI-RRT performs better than ST-RRT* on complex scenes (with more obstacles), has more obstacles on 100\% solved tasks and slower success rate drop. 

%As it can be seen from evaluation, our planner outperforms ST-RRT* and DRGBT in success rate, runtime and arrival time to the goal configuration. 

Figure~\ref{fig:runtime} shows the runtimes of ST-RRT* and SI-RRT on the instances encountering up to 120 obstacles, i.e. the instances for which the SR of ST-RRT* and SI-RRT is 100\% (we did not include DRGBT because of much lower SR). Evidently, as the number of obstacles increases the runtime of the both solvers increases as well. Still SI-RRT scales much better. For 100-120 obstacles it is one order of magnitude faster than SI-RRT* (note that OY axis is in log-scale).

%ST-RRT* has a monotonic increase in running time with increasing number of obstacles. SI-RRT is faster than ST-RRT* and has better asymptotic on large number of obstacles, overtaking it by an order of magnitude in time. 

%A key characteristic of space-time scheduler performance is the arrival time of the robot at the target configuration. In 
Figure~\ref{fig:arrival} shows solution cost of SI-RRT and ST-RRT*, that is a time moment when, according to the constructed plan, the manipulator arrives to the goal configuration. As before the results of DRGBT are omitted due to its low SR. Noteworthy, that ST-RRT* is an anytime planner and has a capability to enhance the cost of the first found solution if time permits. However in our test we do not use this feature for a fairer comparison to SI-RRT which does not use the cost improvement techniques (although they can be encapsulated to the algorithm; we leave this for future work). 
%we have compared ST-RRT* and SI-RRT with respect to this component. We did not include DRGBT in the graph because it has a low success rate and solves only the easy part of the tasks. ST-RRT* and SI-RRT worked up to the first constructed path, without improving it further. 
This explains the poor cost of ST-RRT* for the tests with 1 obstacle. ST-RRT* grows its forward and backward trees from time moment 0 and time moment $t_{max}$, respectively, and when they meet, resulting motion has extremely slow speed. As the number of obstacles increases, this effect diminishes. Still the cost of SI-RRT is better compared to ST-RRT*, that can be attributed to the use of safe intervals and not sampling distinct time moments. 

Overall, the results of the experiments clearly show that, first, the planners that reason over time are superior over the re-active ones (i.e. ST-RRT and SI-RRT* outperform DRGBT) and, second, the suggested solver, ST-RRT, notably surpass ST-RRT* -- current state of the art in high-dimensional path planning in space-time. We believe that the main reason for this is the usage of safe intervals. This confirms that this concept, initially introduced in the context of search-based planning, is beneficail for sampling-based planning as well.

%Overall, comparing SI-RRT with ST-RRT* we conclude that our planner is faster and provides better solutions (compared to the first solutions provided by ST-RRT*). The possible explanations of this may be as follows:
%\begin{enumerate}
    %\item ST-RRT* samples both in configuration space and time. By utilizing safe intervals, SI-RRT decreases the dimensionality of planning that positively influence the runtime;
    %\item ST-RRT* relies pseudo-metric, and uses naive brute-force way to find nearest neighbours. SI-RRT uses $L^2$ norm in configuration space, thus, having ability to use fast nearest neighbours search, such as KD-Trees;
    %\item By using safe intervals, SI-RRT finds narrow passages in space-time faster than ST-RRT*
%\end{enumerate}

%It is also noteworthy that despite DRGBT lags far behind SI-RRT and ST-RRT* it has a decent advantage as it does not need to know the trajectories of the obstacles. Still, generally one may conclude that when the information on scene dynamics is available the suggested planner, SI-RRT, clearly outperforms state of the art.

\section{Conclusion}
In this work we have suggested a novel planner, SI-RRT, tailored to finding safe paths in the predictably changing environments for multi-degree of freedom robotic systems, such as manipulators. SI-RRT leverages a combination of the well-established techniques such as sampling-based planning, bidirectional search and safe interval path planning (for the later the original procedure to compute safe intervals has been proposed). A comprehensive empirical evaluation of SI-RRT comparing it with state of the art was conducted and the planner was shown to notably outperform the competitors.

The prominent directions for future research include: investigating the theoretical properties of SI-RRT, developing anytime variants of SI-RRT that may gradually improve the solution cost (and converge to optimal solutions in the probabilistic sense), utilizing SI-RRT in multi-agent manipulation planning.

\bibliography{main}

\end{document}